# ChatGPT or academic scientist? Distinguishing authorship with over 99% accuracy using off-the-shelf machine learning tools.

Heather Desaire*[1], Aleesa E. Chua[1], Madeline Isom[1], Romana Jarosova[1], David Hua[1]

[1]Department of Chemistry, University of Kansas, Lawrence, Kansas 66045, USA

*Address correspondence to:
Heather Desaire, phone: 785-864-3015, Email: hdesaire@ku.edu

**Summary**

ChatGPT has enabled access to AI-generated writing for the masses, and within just a few months, this product has disrupted the knowledge economy, initiating a culture shift in the way people work, learn, and write.  The need to discriminate human writing from AI is now both critical and urgent, particularly in domains like higher education and academic writing, where AI had not been a significant threat or contributor to authorship. Addressing this need, we developed a method for discriminating text generated by ChatGPT from (human) academic scientists, relying on prevalent and accessible supervised classification methods.  We focused on how a particular group of humans, academic scientists, write differently than ChatGPT, and this targeted approach led to the discovery of new features for discriminating (these) humans from AI; as examples, scientists write long paragraphs and have a penchant for equivocal language, frequently using words like "but", "however", and "although".  With a set of 20 features, including the aforementioned ones and others, we built a model that assigned the author, as human or AI, at well over 99% accuracy, resulting in 20 times fewer misclassified documents compared to the field-leading approach.  This strategy for discriminating a particular set of humans' writing from AI could be further adapted and developed by others with basic skills in supervised classification, enabling access to many highly accurate and targeted models for detecting AI usage in academic writing and beyond.

**Introduction**

ChatGPT, released to the public in November of 2022, has become a media sensation, already attracting over 100 Million subscribers within the first two months.[1] It can offer detailed responses to a wide range prompts, tell jokes, correct grammar in essays, and even write human-sounding research reports. The capabilities of this technology are, at the same time, tantalizing and frightening.  One obvious early concern about the effects of this new platform was its cooption by students to do their homework for them,[2,3] but the concerns soon spiraled up to include its potentially negative impact on academic and professional writing as well.[4]

But how would one go about differentiating ChatGPT-generated text from that produced by a human academic? Researchers have already spent years developing methods to discriminate human-generated from AI-generated text.[5] Numerous options exist, including so called "zero shot" methods, which seek a general solution to the problem without extensive training on particular types of text examples[6,7] by employing statistical methods comparing, for example, words' frequency in a document versus their frequency in the large language model from which the document may have originated.[6] Commonly, deep learning and extensive pretraining is employed instead. One field leader, the RoBERTa detector,[8] replied upon a massive amount 160GB of data for pretraining.[9] This AI detector has shown superior performance in several high-quality studies,[5,10] particularly when it is further tuned on the data of interest.[11] While these prior studies are certainly relevant, AI experts note that ChatGPT's capabilities are "surprisingly strong,"[12] so a careful reassessment of the best way to distinguish this advanced language model from human writing is warranted, particularly in writing destined for the academic literature, which is not well-represented in earlier studies discriminating AI from human text.

One relevant research project that focused on detecting the use of ChatGPT, specifically, studied online data from sources like reddit and Wikipedia, where advice was being provided.[12] The challenge of detecting AI authorship in this type of informal online content was easily met using the RoBERTa detector[8] mentioned above. It typically correctly identified the authorship (human or ChatGPT) at an accuracy of >98% (for complete responses) on a variety of data sets. While this study, with a large body of test data, demonstrates strong proof of principle, we note that many of the key differentiating traits between the humans and ChatGPT--including use of colloquial and emotional language--are not traits that academic scientists typically display in formal writing, so the accuracies seen here would not necessarily translate to academic writing.

A second study that specifically addressed discriminating ChatGPT-derived content from human-generated comparators used data more similar to the data of interest herein.[13] Gao et al. developed a data set of 50 human-generated and 50 ChatGPT-generated abstracts that fit the format of a variety of medical journals; they used several different methods to test their distinguishability. Blinded human reviewers could correctly classify the writing sources less than 70% of the time.[13] The researchers also used an online adoption of RoBERTa, the GPT-2 Output Detector (https://openai-openai-detector.hf.space/), to assess the medical abstracts, but the tool's performance was noticeably weaker than in the aforementioned study[12] assessing informal online content. On medical writing, this AI detector only correctly classified 82% of the abstracts.[13] We note that attributing the reduced classification performance in this case to the *type* of data (scientific abstract) and not some other parameter is premature.

In the work described herein, we sought to achieve two goals; the first is to answer the question about the extent to which a field-leading approach for distinguishing AI- from human-derived text works effectively on discriminating academic science writing as being human-derived or from ChatGPT, and the second goal is to attempt to develop a competitive alternative classification strategy. We focus on the highly accessible online adaption of the RoBERTa model, GPT-2 Output Detector, offered by the developers of ChatGPT, for several reasons. It is a field-leading approach. Its online adaption is easily accessible to the public. It has been well-described in the literature. Finally, it was the winning detection strategy used in the two most similar prior studies (references 12 and 13).

The second project goal, to build a competitive alternative strategy for discriminating scientific academic writing, has several additional criteria. We sought to develop an approach that relies on (1) a newly developed, relevant data set for training, (2) a minimal set of human-identified features, and (3) a strategy that does not require deep learning for model training, but instead focuses on identify writing idiosyncrasies of this unique group of humans, academic scientists.  Applying classical machine learning methods to a set of linguistic features for discriminating human- and AI- text has been successfully applied to distinguish some language models from human writing.[11] The challenging part is to find the appropriate feature set for ChatGPT, relevant to academic science writing. If this route for distinguishing human-derived text from ChatGPT is demonstrated to be a viable possibility, this precedent could expand the opportunities for quick, focused solutions for discriminating human from AI text in targeted knowledge domains, such as academic writing in general, plagiarism in particular courses, and beyond. Furthermore, the approach could be adopted and further developed by researchers without deep learning expertise, significantly expanding the number of individuals capable of research in this critically important and emerging domain.  We attempted to design a competitive detection strategy using off-the-shelf machine learning tools that would be at least as good as the RobERTa detector; surprisingly, we succeeded.

**Results**

As one key goal of this project was to develop an original method for differentiating human-generated academic writing from text generated by ChatGPT, we first asked the question:  What documents should comprise the data set?  What prompt(s) should ChatGPT receive, and what human-derived text should it be compared to?  While many strategies would be reasonable, we decided to use Perspectives articles from the journal, Science, for the human-derived text and to pair each example with text generated from ChatGPT, based on both the exact title from the Science article and a human-derived title that maximally captured the topics addressed in the comparator text.  This approach had the following merits:  The writings would cover a diverse range of topics, from biology to physics, so the resulting model would not be biased by a particular disciplines' mores and vocabulary, thus potentially making the resulting feature set more broadly useful. Second, the Perspectives documents are written by scientists, not journalists, so they would reflect scientists' writing and not those who write *about* science. Finally, these articles typically describe a research advance present in the issue in which they appear (although some cover other topics, like remembrances of scientists who recently died), so their content is highly similar to the type of content that ChatGPT produces; as a large language model, ChatGPT can summarize recent scientific developments or the contributions of a famous person, but it does not describe experiments or findings for the first time. A sufficiently diverse data set, with enough examples, was desirable, so the model would not be influenced by a few authors' writing style or vocabulary.

     We chose 64 Perspectives articles from September, 2022, through March, 2023, and used them to generate 128 ChatGPT examples for the training set. This group of training data generated 1276 example paragraphs for the training set, each paragraph considered a sample. (ChatGPT-generated training data and an inclusive list of the human-generated training data are provided in Supplemental Data.) After the model was fully developed and optimized, we also generated two test sets, from November/December, 2021, and November/December, 2020. Each test set had 30 Perspectives articles from humans and 60 ChatGPT-derived essays, generated as described above. A total of 1210 example paragraphs populated the test sets, providing enough data to get meaningful statistics at both the paragraph-level and the document level. Approximately 60% of the paragraphs were from ChatGPT, for both the training and test data sets.

*Feature Development.*

Through manual comparison of many examples in the training set, we identified four categories of features that appeared to be useful in distinguishing the human writing from the chatbot. They are: (1) paragraph complexity; (2) sentence-level diversity in length; (3) differential usage of punctuation marks; (4) different "popular words". While these feature categories were arrived at independently, through text comparisons, we note that three of the four categories are similar to feature types used elsewhere. Sentence-length diversity is one feature that is also used in an online AI detector (GPTZero; https://gptzero.me/), although the underlying model, method, and relative effectiveness of that tool has not been publicly disclosed or peer reviewed. Differential use of punctuation has been reported previously;[12] although in that case, the primary example was that of humans using multiple exclamation points (!!) to add emphasis, a practice not used in academic writing. Finally, linguistic features, and in particular, commonly used words, have previously shown utility in identifying AI writing,[6] but in the prior example, the entire corpus of texts used to train the large language model was used for statistical assessment of the writing samples. We are unaware of any reports of researchers noting the utility of paragraph complexity as a feature type; perhaps that is because scientists, as a group, write more complex paragraphs than the human writings from existing data sets. Furthermore, we are unaware of any prior research combining this group of feature types to build a model. We ultimately identified 20 features that fit into one of these four categories and could be potentially useful in the classification task. Table 1 shows each feature that was included in the final model, the general category to which it belongs, and the prevailing difference detected between the human- and AI-generated training data.

**Table 1: Features in the Model**

| Feature Number | Feature Type (1-4)[a] | Short Description | Greater In: |
|---|---|---|---|
| 1 | 1 | Sentences Per Paragraph | Human |
| 2 | 1 | Words Per Paragraph | Human |
| 3 | 2 | " ) " present | Human |
| 4 | 2 | " - " present | Human |
| 5 | 2 | " ; " or " : " present | Human |
| 6 | 2 | " ? " present | Human |
| 7 | 2 | " ' " present | ChatGPT |
| 8 | 3 | Standard deviation in sentence length | Human |
| 9 | 3 | Length difference for consecutive sentences | Human |
| 10 | 3 | Sentence with < 11 words | Human |
| 11 | 3 | Sentence with > 34 words | Human |
| 12 | 4 | Contains "although" | Human |
| 13 | 4 | Contains "However" | Human |
| 14 | 4 | Contains "but" | Human |
| 15 | 4 | Contains "because" | Human |
| 16 | 4 | Contains "this" | Human |
| 17 | 4 | Contains "others" or "researchers" | ChatGPT |
| 18 | 4 | Contains numbers | Human |
| 19 | 4 | Contains 2 times more capitals than " . " | Human |
| 20 | 4 | Contains "et" | Human |

[a.] Feature types: 1: Paragraph complexity. 2: Punctuation marks. 3: Diversity in sentence length. 4: Popular words or numbers.

*Ways in which ChatGPT produces less complex content than human scientists.* Two of the four categories of features used in the model are ways in which ChatGPT produces less complex content than humans. The largest distinguishing features were the number of sentences per paragraph and the number of total words per paragraph. In both cases, ChatGPT's averages were significantly lower than the human scientists. We also found that humans preferred to vary their sentence structures more than ChatGPT: while the *average* sentence length was not a useful discriminator of the two groups, the *standard deviation* of the sentence length, in any given paragraph, was a valuable differentiator, as was the median difference (in words) between a given sentence and the one immediately following it. Humans vary their sentence lengths more than ChatGPT. Humans also more frequently used longer sentences (35 words or more) and shorter sentences (10 words or fewer.)

*Ways in which ChatGPT writes stylistically differently than human scientists.*  The remaining two categories of differentiating features could be described more as "stylistic" choices.  On one hand, human scientists more frequently use question marks, dashes, parentheses, semicolons, and colons; while ChatGPT uses more single quotes.  Scientists also use more proper nouns and/or acronyms, both of which are captured in the frequency of capital letters, and scientists use more numbers.  ChatGPT seems to prefer to be more general with the information it provides, and this overriding theme shows up in differences in specific word frequencies.  ChatGPT is more likely to refer to ambiguous groups of people, including "others" and "researchers", while humans are more likely to name the scientist whose work they are describing.  Human scientists also displayed other consistent patterns in the training data:  they are more likely to use equivocal language (however, but, although); they also use "this" and "because" more frequently.

*Classification of training data.*  Before ultimately assessing the utility of the set of features in discriminating ChatGPT from humans, several off-the-shelf classifiers were initially auditioned, and XGBoost was selected as the classifier that provided superior performance. The performance was tested using "Leave-One-Essay-Out" cross-validation.  (In this case, every paragraph from a particular essay being tested would be left out of the model, and the remaining 1200+ paragraphs would be used to build a model to classify the omitted examples.)  This approach was chosen as a more rigorous option instead of a LOOCV method, where only a single paragraph would be removed, or, for example, a ten-fold cross-validation, where 10% of the data would be randomly removed. The chosen method removed the possibility that any examples from a given writer, or an individual essay, in the case of ChatGPT, could be used to make it easier to identify other paragraphs within the same essay (by the same writer).  The model, therefore, does not rely on having any previous examples of the human writer's works whose essay is being classified.

The results, shown in Table 2, indicate that the chosen feature set and classifier are effective at discriminating human writing from ChatGPT. At the level of an individual paragraph, 94% of the >1200 examples are correctly classified. Since every essay contained multiple paragraphs, we also classified each document based on which class was assigned to the majority of the paragraphs. In cases where there are an equal number of paragraphs assigned to each class, the overall assignment was made based on the first paragraph only. At the document level, 99.5% of the samples are correctly assigned. Only one of 192 is misassigned. The single misclassified document is a description of a recently deceased scientist and his works, not an article describing a scientific advance. Finally, we also tracked the accuracy of the assignments based on the first paragraph only and found that making a classification based on all the paragraphs is better than using just a single paragraph, which is not surprising. However, if just one paragraph could be tested, choosing the first one gives slightly better results than a randomly selected paragraph. With this high level of performance, we decided to move on to testing the model on newly acquired data; two test sets were acquired.

**Table 2: Model Accuracy in Training and Test Sets**

|          | Paragraph-level Statistics | | | Document-level Accuracy | |
|----------|----------|----------|-------|---------|---------------|
|          | Examples | Accuracy | AUC   | Overall | 1st Paragraph |
| Training | 1276     | 94%      | 0.934 | 99.5%   | 95%           |
| Test 1   | 614      | 92%      | 0.913 | 100%    | 99%           |
| Test 2   | 596      | 92%      | 0.914 | 100%    | 97%           |

The test data were also Perspectives articles from Science, but they were from slightly older issues (2021 and 2020, instead of 2022/2023.) The same features and model were used, with all the training data now leveraged to build a single model to assess the examples in the test sets. While a slight, and expected, dip in accuracy occurs at the paragraph level, we were pleased to see that at the document level, the model is 100% accurate for the 180 examples in the (combined) test sets. Furthermore, the model's accuracy for the first paragraph of each document is 97% in one data set and 99% in the other, further supporting the finding that testing a first paragraph provides better results than testing any random paragraph. In summary, this approach exceeds our expectations, particularly at the document level, although additional effort in identifying more or better features may result in improved performance when the goal is to assign the ownership of a single paragraph.

The results of the newly built model for detecting writing from ChatGPTcan be best appreciated when contextualized against an existing state-of-the-art method. For the reasons described in the introduction, the method of choice for this comparison was the online-accessible version of the RoBERTa detector, GPT-2 Output Detector. Each paragraph of text from the training and test sets was provided to the Output detector, one at a time, and the authorship (human or AI) was assigned based on the output score, which ranges from 0 to 100% for both groups. The scores were converted to binary assignments (human or AI), and the results for both the training and test sets were tallied, both at a single paragraph level and at a full document level. The results are shown in Table 3. The GPT-2 Output Detector was inferior to the method described herein for every assessment conducted. Most notably, at the full document level, which comprised at least 300 words from ChatGPT and typically more than 400 words for the human examples, the Output Detector misassigned 20 documents, while the method described herein misassigned just one in a total of 372 examples.

**Table 3: Accuracy of GPT-2 Output Detector**

|  | Paragraph Accuracy. | Document Accuracy | # Documents misassigned |
|---|---|---|---|
| Training Set | 86% | 96% | 8 |
| Test Set 1 | 88% | 92% | 7 |
| Test Set 2 | 85% | 94% | 5 |

**Discussion**

This study is the first to demonstrate a highly effective approach for differentiating human-generated academic science writing from content produced by ChatGPT. Since academic scientific writing is fundamentally different in style than most online content, like restaurant reviews or informal communication on discussion boards, new and different ways to differentiate the text were considered.  Some of the derived features, such as those indicating humans' preference for more diversity in their sentence lengths, have been previously touted as useful (and referred to as "burstiness"), although they are not sufficient on their own to effectively identify the author as human or AI. Other features, like scientists' penchant for writing long paragraphs and using equivocal language like "however", "but", and "although", were newly identified.  These features may not be useful in authenticating a human author for informal writing examples, but they are likely useful for a variety of academic writings.  We consider, then, the main contribution of this work to be a *method,* not *a universal model.*  By considering these four categories of features, together namely: (A) paragraph length, (B) diversity in sentence length, (C) punctuation differences, and (D) popular words among a given group, effective models for discriminating human-derived from chatbot-derived text are likely achievable in many domains, particularly in academic and scholarly literature. We further note that the success herein was enabled at least in part by focusing on the idiosyncrasies of academic scientists.  It is possible that this approach could be broadly applicable for assessing academic writings as being from humans or AI, by simply substituting in different "popular words", used by the subset of humans for which the model is being developed.

    In extending this strategy to other or new circumstances, several additional options for distinguishing documents could be used in conjunction with or in lieu of the ones described here.  First, the number of popular words or word types that vary between the training and test groups could be expanded; the list presented here does not comprehensively capture the differences between these two types of text, and other text types could certainly have other key words that discriminate them.  Furthermore, an approach similar to the one used in Gehrmann *et al.* could become a component of the classification strategy;  they assigned each word in the

document a score for being commonly or uncommonly used in the language model.[6] Another possible avenue for differentiating documents longer than a paragraph includes identifying useful document-level features. In this work, we used no document-level features, but they may be valuable in other cases. For example, the diversity in paragraph length is larger in human-generated text than ChatGPT. In fact, simply using the standard deviation of the number of words in each paragraph throughout a given document produces a highly predictive indicator of whether the document's author is human. The AUC for this single feature is 0.98 for the training data. While this single calculation is simple, quick, and surprisingly accurate, it was not utilized in this work because the strategy of assigning a class to each paragraph, and making the final assignment based on the class with the most assigned paragraphs, produced more accurate results. We note that using this single feature, standard deviation of paragraph length, is more predictive than the GPT-2 Output detector's assignments on full documents. In cases where paragraph-level differences are difficult to detect, using document-level features, particularly ones assessing diversity in paragraph lengths, may increase the discriminating capacity of the model.

    In this work we achieved two objectives, to assess the extent to which an existing field-leading tool (GPT-2 Output Detector) could differentiate from human- or AI-derived writing in the context of academic science, and to identify a strategy with a competitive advantage. While the RoBERTa algorithm used in the Output Detector is effective for discriminating ownership of online content, where language is informal and humans are emotive,[12] its performance sags considerably on the classification task described herein. We note that the Output Detector's performance at the paragraph level (85% to 88%) is not considerably different than the 82% accuracy previously reported on 100 scientific abstracts,[13] and this may be generally representative of the accuracy achievable from this device on a single paragraph of scientific academic text. The new method and model described herein was more effective at the

paragraph level and at *much* more effective at the document level, with twenty times fewer errors on full-document assignments. We note, though, that our approach was designed to be applicable to a narrower scope of writing, and the extent to which the model is broadly translatable is yet to be determined. More likely than having developed a new universal model for identifying the authenticity of academic writing from humans, the general approach, of using these four feature types (substituting in a specific field's "popular words") and supervised classification on a reasonable set of highly representative training data, will likely produce superior classification results for academic writing compared to using an untuned, general purposed classifier, even one whose original training required deep learning and hundreds of gigabytes of data.

**Experimental Details**

*Data set development.*

All content from ChatGPT was generated between March 2, 2023, and March 17, 2023. A typical prompt for ChatGPT would say: "Can you produce a 300 to 400 word summary on this topic: A surprising fossil vertebrate". An excel file containing the complete list of prompts provided to ChatGPT is included in the Supplementary Data. The resulting text varied in both the number of paragraphs and the total number of words, but no effort was made to control for these apects beyond the initial instruction. The complete set of ChatGPT-derived content is available in the Supplemental Data section as well. Note that each row in the matrix contains a single ChatGPT paragraph. A key linking these paragraphs to their prompt is provided as the first column of the matrix.

The human-generated content was extracted from Perspectives articles in Science. For the training set, the complete set of Perspectives from September, 2022, was used (17 articles). The first 30 Perspectives, starting from the November 4 (2022) issue and continuing through the December 9 (2022) issue, were also selected. Finally, 17 Perspectives articles from February and March, 2023, were selected. No restrictions based on content or length were imposed on the selected articles. After selection, the text from the body of the article was copied into a text file, assuring that the paragraph delineation matched the original documents' delineations. Any figures, tables, legends, or references were removed. Finally, if in-text citations were present, they were deleted, along with the parentheses that enclosed them. No other modifications were made. As with the ChatGPT data, each paragraph of text occupied a single row of a data matrix, and each row became a unique "sample" in the data set.

*Feature Extraction and classification.*

A total of 20 features were extracted from the textual data to comprise the training set and test set. Example code used for each feature's extraction is provided in Supplemental Data. All data analysis was done in RStudio, using R version 4.0.3. After the matrix of samples and features was built, its utility for discriminating the author type (AI or human) was initially tested using several off-the-shelf classifiers, including XGBoost (using the package xgboost), Naïve Bayes (using the package e1071), and AC.2021[14] (using the code provided in the cited manuscript). Since XGBoost produced lower error rates on the training data, it was chosen as the classifier for the remainder of the work. The parameter set from XGBoost included the following:

```
params <- list(booster = "gbtree", objective = "multi:softmax", num_class=2, eta=0.2, gamma=0, max_depth=6, min_child_weight=1, subsample=1, colsample_bytree=1)
 xgb1 <- xgb.train (params = params, data = dtrain, nrounds = 50, maximize = F )
```

The accuracy of the model on training data was assessed by using "leave-one-essay-out" cross-validation. In this paradigm, a single essay at a time, of the 192 essays comprising the training data, was left out of the model, and the trained model was used to classify all the remaining paragraphs in the data set that originated from the left-out essay. After 192 rounds of this, every paragraph in the training set was classified. The class assignment, at the document level, was made by a voting strategy, where each paragraph's assignment was considered one vote, and the class with the highest number of votes was assigned to the document during document-level classification. In cases where both classes received equal number of votes, the assignment was based on the class assigned to the first paragraph. Accuracy statistics were based on the number of correct assignments vs the number of total assignments. AUC statistics were calculated using the package, pROC. Test data was assigned in an identical matter to the training data, with the exception that all 1276 paragraphs in the training data were used to build a single model, and that single model was used for both test sets described in the manuscript.

**Supplemental Information.** The following additional documents and tools are provided: A complete list of references and URLs for the human-produced text "ListOfDocsAndPromptsForSuppData.xls"; the aforementioned document also contains the complete list of prompts provided to ChatGPT; a matrix with all the text examples generated from ChatGPT "Mat.csv", a key linking each paragraph of the ChatGPT text (each line in the data matrix) to the writing example and prompt is the first column in the "Mat" matrix; example code used to extract the features utilized in the model "ExampleFeatureBuilder.txt".

**Acknowledgements.** This work was supported by NIH grant R35G130354 (to HD) and funding from the Madison and Lila Self Graduate Fellowship, University of Kansas (to EC and MI).

**Author Contributions.** Conceptualization and methodology: HD and DH; data acquisition: AC, MI, RJ; model development: HD, DH, AC, testing: HD, AC, MI, RJ; manuscript writing: HD. All authors have read and approved the final version of the manuscript.